# Transformer-based Automatic Post-Editing with a Context-Aware Encoding Approach for Multi-Source Inputs

**WonKee Lee, Junsu Park, Byung-Hyun Go, Jong-Hyeok Lee**
Department of Computer Science and Engineering,
Pohang University of Science and Technology (POSTECH), Republic of Korea
{wklee, jspak3, briango28, jhlee}@postech.ac.kr

## Abstract

Recent approaches to the Automatic Post-Editing (APE) research have shown that better results are obtained by multi-source models, which jointly encode both source (*src*) and machine translation output (*mt*) to produce post-edited sentence (*pe*). Along this trend, we present a new multi-source APE model based on the Transformer. To construct effective joint representations, our model internally learns to incorporate *src* context into *mt* representation. With this approach, we achieve a significant improvement over baseline systems, as well as the state-of-the-art multi-source APE model. Moreover, to demonstrate the capability of our model to incorporate *src* context, we show that the word alignment of the unknown MT system is successfully captured in our encoding results.

## 1 Introduction

Thanks to recent advances in deep learning over the last two decades, machine translation (MT) has shown steady improvements (Bahdanau et al., 2014; Gehring et al., 2017; Vaswani et al., 2017). Nevertheless, even the state-of-the-art MT systems are still often far from perfect. Translations provided by MT systems may contain incorrect lexical choices or word ordering. Therefore, to get publishable quality translations, correction of MT errors has been regarded as an essential post-processing task referred to as "post-editing" (PE).

From the MT point of view, PE can be defined as the process of editing translations provided by MT systems with a minimal amount of manual effort (TAUS Report, 2010) while the field of "Automatic post-editing" (APE) aims to mimic the human post-editing process automatically. Given the underlying assumption of the APE task in which the source text (*src*), the corresponding output of the unknown MT system (*mt*), and its human post-edited sentences (*pe*) are the only available resources, many studies have attempted to leverage information from both *src* and *mt* by suggesting multi-source architectures (Chatterjee et al., 2016; Libovický et al., 2016). Furthermore, as the newly proposed architecture called the Transformer network (Vaswani et al., 2017) has shown to be effective in various sequence-to-sequence problems, various multi-source adaptations of the Transformer (Junczys-Dowmunt and Grundkiewicz, 2018; Shin and Lee, 2018; Tebbifakhr et al., 2018) have been applied to APE.

In this work, we propose a new multi-source APE model by extending Transformer. We focus especially on modifying its original encoder portion to construct effective joint representations of two sources (*src*, *mt*). A major advantage of our approach is that by constructing the joint representation such that *src* context is incorporated into *mt* representation, the model allows information on *mt* as well as its *src* context to be considered together in generating *pe*. In addition, we utilize a weight sharing method between the embedding and output layers in a manner similar to Press and Wolf (2017), and Junczys-Dowmunt and Grundkiewicz (2018). Finally, our model outperforms not only baseline systems but also other multi-source APE models.

## 2 Related Work

Properly modeling the relations between multiple sources is important for solving multi-source sequence generation problems such as APE. For this purpose, Chatterjee et al. (2016) suggested the Factored APE model, combining *mt*, its *src* alignment, and other features into factors. It is based on a statistical model, but the approach to incorporating *src* context into *mt* is similar to our work.

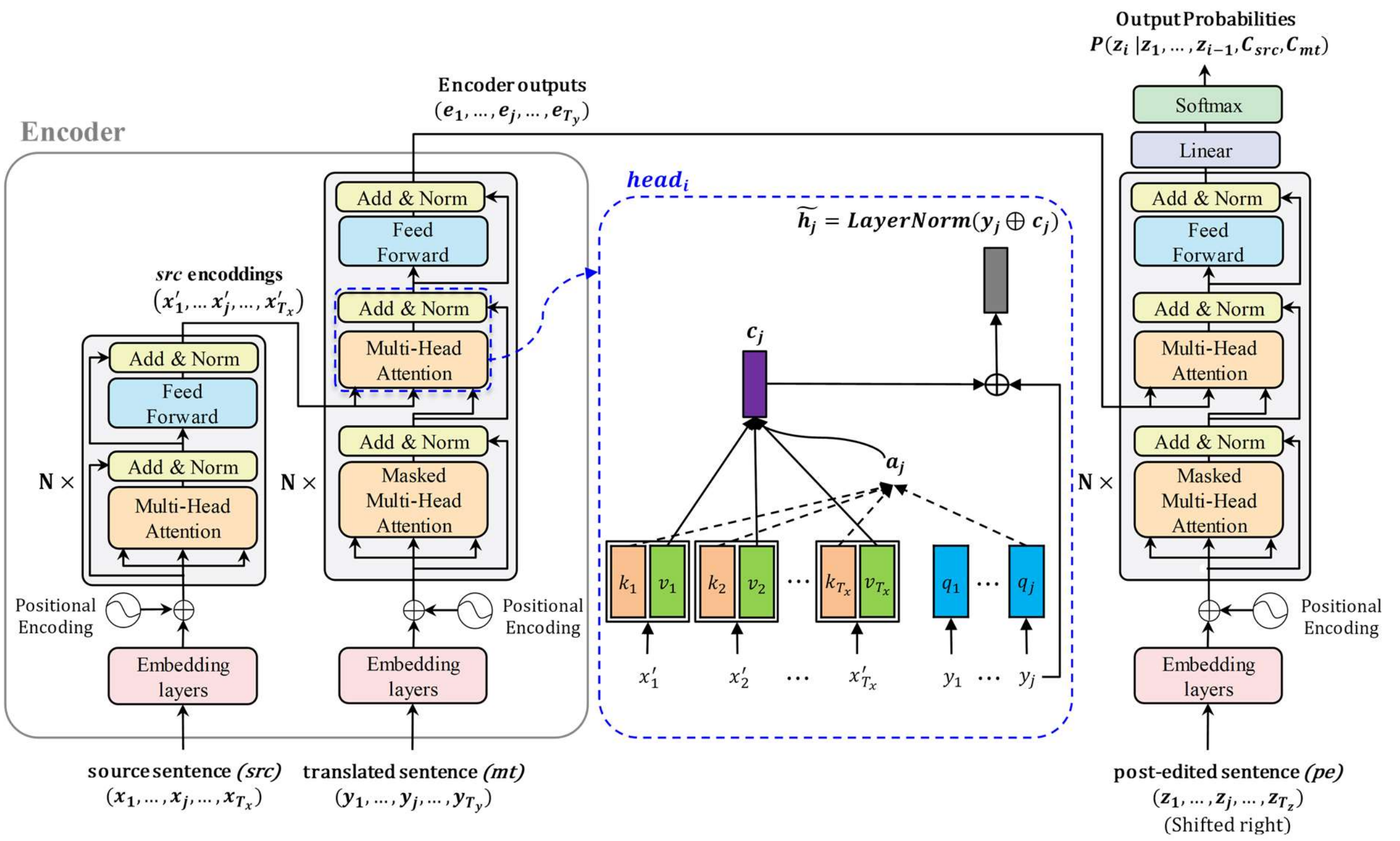

Figure 1: **An illustration of the proposed APE architecture** – a square with dashed line shows joint hidden representation at j[th] position with respect to *src* context and previous *mt* context before j.

For neural network-based approaches, Libovický et al. (2016) used an extended RNN encoder-decoder architecture (Bahdanau et al., 2014) with two separate encoders for each source so that the decoder could attend to both. Chatterjee et al. (2017) expanded on this model to win the WMT 17 APE shared task. Recently, a variety of multi-source architectures based on the Transformer (Junczys-Dowmunt and Grundkiewicz, 2018; Shin and Lee, 2018; Tebbifakhr et al., 2018) have been suggested. They are all architectures employing separate encoders but with slightly different ways of modeling the joint representation: Tebbifakhr et al. (2018) simply concatenated the output of two encoders; Shin and Lee (2018) considered multiple relations (*mt-pe*, *src-pe*, *src-mt*); Junczys-Dowmunt and Grundkiewicz (2018) encoded src and *mt* using shared weight parameters. The key difference between their works and ours is that they encode two sources independently, while our method feeds *src* encoding results as inputs when encoding *mt*, so that src context can be directly integrated into each *mt* representation.

## 3 Multi-Source Automatic Post-Editing

We predicted that considering *mt* with its corresponding *src* context leads to an improvement in post-editing performance. The advantage of this approach is made clear in cases for which *src* context must be taken into consideration for accurate post-editing. For example, the German phrase "mein Haus" ("my house" in English) is a possible mistranslation of the English phrase "my home". A system that recognizes *src* context would be able to correct it into "mein Zuhaus". In the following, we describe how we expand on this intuition in detail.

### 3.1 Model Architecture

Given the sequences *src* $\mathrm{x} = (x_1, \ldots, x_{T_x})$, *mt* $\mathrm{y} = (y_1, \ldots, y_{T_y})$, and *pe* $\mathrm{z} = (\mathrm{z}_1, \ldots, z_{T_z})$ with the length of the sequence $T_x, T_y$, and $T_z$, respectively, the model is trained to learn probabilities conditioning on two sources x, y and target history $z_{<n}$:

$$P(\mathrm{z}|\mathrm{x}, \mathrm{y}; \theta) = \prod_{n=1}^{N} P(z_n|\mathrm{x}, \mathrm{y}, z_{<n}; \theta) \qquad (1)$$

The model architecture, as shown in Figure 1, is an extension of Transformer. As described in Vaswani et al. (2017), each stacked layer is composed of multi-head attention networks and position-wise fully connected feed-forward networks. Unlike the original Transformer, to construct joint representations for two inputs, the encoder consists of two sub networks: ***src encoder*** $enc_{\theta_{src}}(\mathrm{x})$ and ***mt encoder*** $enc_{\theta_{mt}}(\mathrm{x}', \mathrm{y})$, shown on the left and right side of the encoder in Figure 1. The workflow of the model is as follows:

- $enc_{\theta_{src}}(\mathrm{x})$ accepts *src* embeddings as input and produces a sequence of encoded vectors $\mathrm{x}' = (x'_1, \dots, x'_{T_x})$ in which each *src* is encoded with its context via the self-attention layer.

- $enc_{\theta_{mt}}(\mathrm{x}', \mathrm{y})$ accepts both *mt* embeddings and the output of $enc_{\theta_{src}}(\mathrm{x})$ as inputs and returns the final output of the encoder $\mathrm{e} = (e_1, \dots, e_{T_y})$ in which each *mt* is jointly encoded with its corresponding *src* context through the attention layer.

- Finally, the decoder, shown on the far right of Figure 1, generates an output sequence z which maximizes (1) at each step by attending to relevant parts of the output of the encoder in an auto-regressive manner (Graves, 2013).

In addition to shared embedding described in Vaswani et al. (2017), we also utilize weight sharing across the embedding and output layers in a manner similar to Junczys-Dowmunt and Grundkiewicz (2018).

### 3.2 Multi-Head Attention Layer

The multi-head attention layers marked with dashed lines in Figure 1 play an important role in constructing joint representations. As described in Vaswani et al. (2017), we utilize the same multi-head attention with *h*-heads based on scaled dot-product attention to get matrix *C* composed of context vectors as follows:

$$C = \mathrm{MultiHead}(Q, K, V) = \mathrm{Concat}(head_1, \dots, head_h)W^O \quad (2)$$

where

$$head_i = \mathrm{Attn}(QW_i^Q, KW_i^K, VW_i^V) \quad (3)$$

$$\mathrm{Attn}(Q, K, V) = \mathrm{softmax}\left(\frac{QK^T}{\sqrt{d_k}}\right)V \quad (4)$$

the output is then transferred to the Layer Normalization layer with residual connection. Thus, the output hidden states can cover the context information of *V* along with the hidden state itself as follows:

$$\widetilde{H} = \mathrm{LayerNorm}(H + \mathrm{MultiHead}(Q, K, V)) \quad (5)$$

For $enc_{\theta_{src}}(\mathrm{x})$, *src* embeddings $\mathrm{x} \in \mathbb{R}^{T_x \times dim}$ are assigned to *Q*, *K* and *V*, resulting in hidden states that contain self-context and *src* itself. $enc_{\theta_{mt}}(\mathrm{x}', \mathrm{y})$ consists of two multi-head attention layers. As the first layer works similar to $enc_{\theta_{src}}(\mathrm{x})$ except that future mask is added to mimic the decoding process of the MT system in which predictions can depend only on past information, thus *mt* embeddings $\mathrm{y} \in \mathbb{R}^{T_y \times dim}$ are assigned to *Q*, *K* and *V*. In the second multi-head attention layer, *mt* embeddings are assigned to $Q \in \mathbb{R}^{T_y \times dim}$ and the outputs of $enc_{\theta_{mt}}(\mathrm{x}', \mathrm{y})$ to $K \in \mathbb{R}^{T_x \times dim}$ and $V \in \mathbb{R}^{T_x \times dim}$, resulting in joint representations that contain information of *mt* with its corresponding *src* context. The decoder structurally identical to $enc_{\theta_{mt}}(\mathrm{x}', \mathrm{y})$ therefore predicts a *pe* word depending on both previously generated *pe* words and the final output of the encoder that contains contextual information from both *src* and *mt*.

## 4 Experiments

### 4.1 Data

We trained an English-to-German APE system with the WMT dataset (Bojar et al., 2017) from the IT domain, consisting of 23K and 1K triplets (*src*, *mt*, *pe*) for training and development set, respectively. In addition, we adopted a large sized artificial dataset (Junczys-Dowmunt and Grundkiewicz, 2016), which contains ~4M triplets generated from a round-trip translation, to prevent early overfitting from the small size of the WMT training data. Furthermore, we encoded every sentence into subword units (Kudo, 2018) with a 32K shared vocabulary.

### 4.2 Training Details

We employed the OpenNMT-py (Klein et al., 2017) framework to modify the original Transformer network. We trained our model for ~14K update steps with the Adam optimizer (Kingma and Ba, 2014), warm up learning rates (Vaswani et al., 2017) with a size of 12,000, and batch size of approximately 17,000 tokens for each triplet. Other detailed settings were the same asVaswani et al. (2017). To obtain a single trained model, it consumed ~14K update steps until convergence on the development set.

### 4.3 Evaluation

For evaluation, we tested on the WMT16 and WMT17 APE test datasets with case sensitive TER

| Systems | En-De test WMT 16 | | En-De test WMT 17 | |
|---|---|---|---|---|
| | **TER ↓** | **BLEU ↑** | **TER ↓** | **BLEU ↑** |
| WMT baseline | 24.76 | 62.11 | 24.48 | 62.49 |
| Single-source Transformer baseline (*src→pe*) | 23.82 | 64.01 | 26.93 | 59.78 |
| Single-source Transformer baseline (*mt→pe*) | 20.52 | 69.95 | 20.85 | 69.18 |
| Chatterjee et al. (2017) (Ensemble) | 19.32 | 70.88 | 19.60 | 70.07 |
| Shin and Lee (2018) (Ensemble) | 19.15 | 70.88 | 18.82 | 70.86 |
| Junczys-Dowmunt and Grundkiewicz (2018) (Ensemble) | 18.86 | 71.04 | 19.03 | 70.46 |
| Proposed model (Single model) | **17.80*** | **72.87*** | **18.13*** | **71.80*** |

Table 1: **Evaluation results** – **Bold** refers to our final proposed model. '*' indicates that the improvement over all baselines are statistically significant ($p < 0.01$).

and BLEU scores. For the baselines, we employed 1) the raw MT outputs provided by the test dataset, which serves as the official baseline in WMT, and 2) the original single-source Transformer model trained on *src→pe* and 3) on *mt→pe*. For comparison against other APE systems, we selected three models proposed by 1) Chatterjee et al. (2017), 2) Junczys-Dowmunt and Grundkiewicz (2018), and 3) Shin and Lee (2018), which are the recent multi-source approaches (§2) experimented on the same amount of training data as ours. As shown in Table 1, our model performed significantly better than all baselines. Moreover, we exceeded not only the RNN based multi-source APE system (Chatterjee et al., 2017), which was the winning entry in WMT17, but also other recently proposed transformer based multi-source APE systems which have shown notable results, including the winning entry (Junczys-Dowmunt and Grundkiewicz, 2018) in WMT18.

### 4.4 Analysis

As word-alignment for *mt* sentences can be obtained from the WMT Quality Estimation task (Bojar et al., 2016), we analyzed the similarity of the attention from our encoder and the word-alignment from the unknown MT system to determine if context information in reasonably integrated into the joint representations.

Given the randomly sampled sentence from the WMT 16 development data in Figure 2, we observe that our attention results indicate similar tendency to word-alignment from the MT system. Consequently, we believe that the output of the encoder may successfully incorporate information of *mt* and its corresponding *src* context leading to performance improvement. We have described comparisons for various sentences in the appendix.

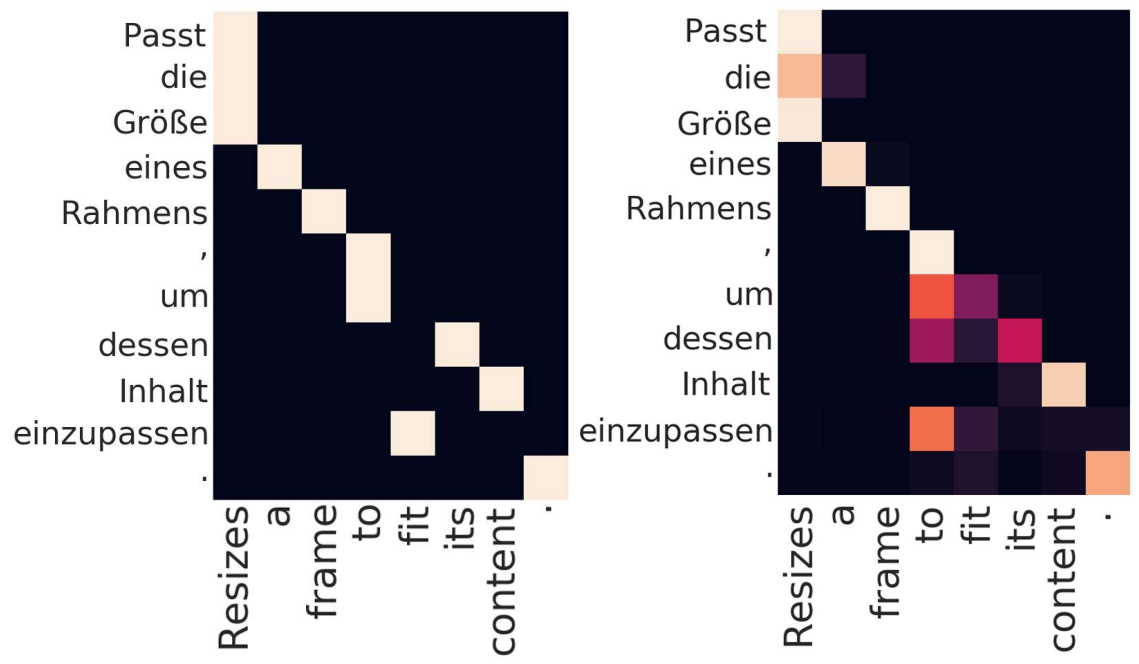

Figure 2: *src-mt* alignment from a random sample. The x and y-axis of each plot correspond to *src* and *mt*, respectively. Each cell shows the alignment probability. The lighter the color, the higher the probability. (a) and (b) refer to the unknown MT system and our encoder, respectively.

## 5 Conclusion

We proposed a new multi-source APE model based on the Transformer network. With the motivation that jointly representing *mt* with its *src* context might be useful in post editing, we have constructed joint representations in our modified encoder. Particularly, that we were able to generate similar alignments to the MT system is, according to our analysis, an indication of the effectiveness of our model in capturing the characteristics of unknown MT systems.

## Appendix

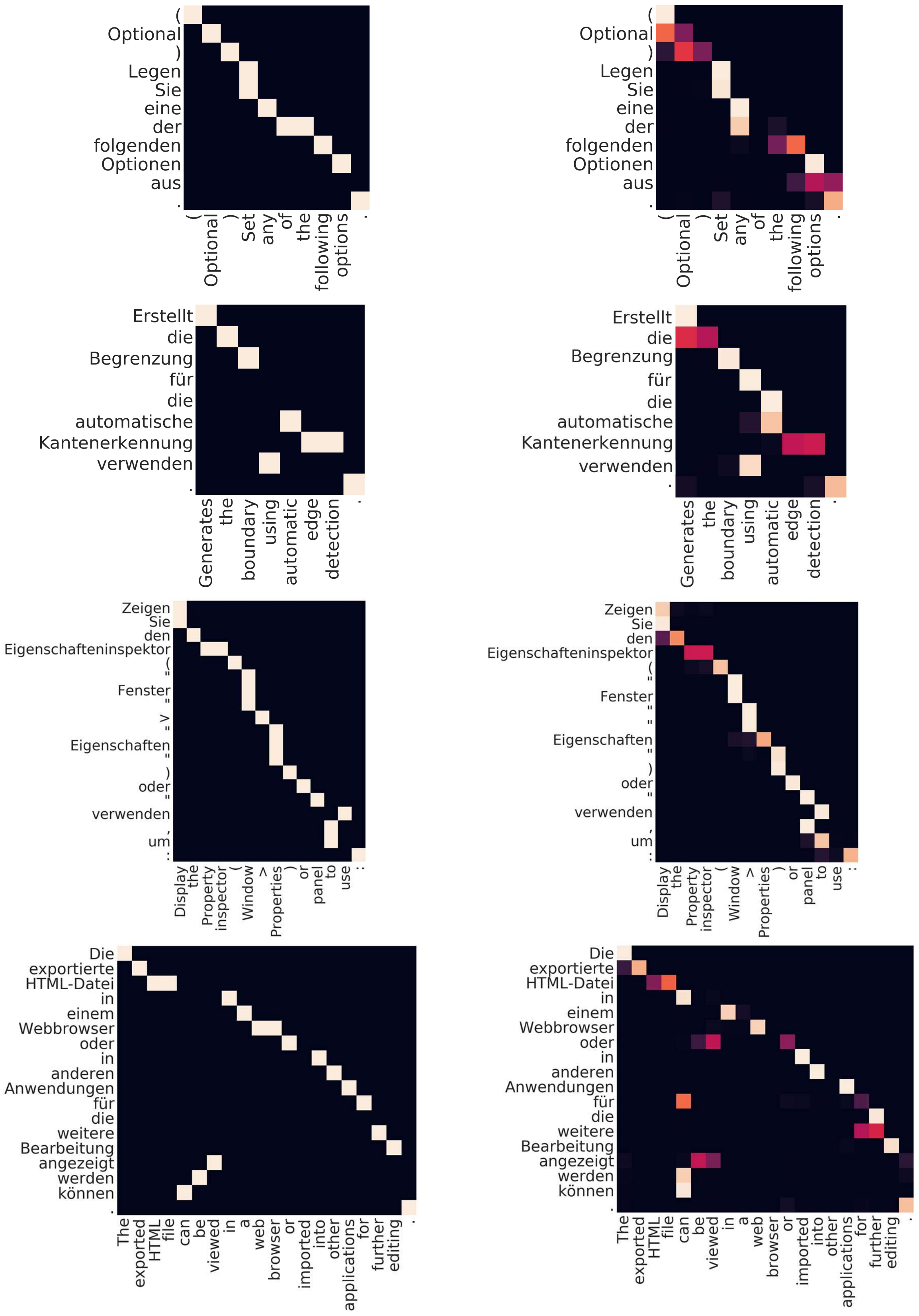

Figure 3: The two columns each correspond to the unknown MT alignments and attention results of our encoder, respectively. The four rows each correspond to the results on short and simple sentences, short and complex sentences, long and simple sentences, long and complex sentences, respectively.